# Sparse-posterior Gaussian Processes for general likelihoods

**Yuan (Alan) Qi**
CS & Statistics Departments
Purdue University
West Lafayette, IN 47906
alanqi@cs.purdue.edu

**Ahmed H. Abdel-Gawad**
ECE Department
Purdue University
West Lafayette, IN 47906
aabdelga@purdue.edu

**Thomas P. Minka**
Microsoft Research
7 J J Thomson Avenue, Cambridge
CB3 0FB, UK
minka@microsoft.com

## Abstract

Gaussian processes (GPs) provide a probabilistic nonparametric representation of functions in regression, classification, and other problems. Unfortunately, exact learning with GPs is intractable for large datasets. A variety of approximate GP methods have been proposed that essentially map the large dataset into a small set of basis points. Among them, two state-of-the-art methods are sparse pseudo-input Gaussian process (SPGP) (Snelson and Ghahramani, 2006) and variable-sigma GP (VSGP) Walder et al. (2008), which generalizes SPGP and allows each basis point to have its own length scale. However, VSGP was only derived for regression. In this paper, we propose a new sparse GP framework that uses expectation propagation to directly approximate general GP likelihoods using a sparse and smooth basis. It includes both SPGP and VSGP for regression as special cases. Plus as an EP algorithm, it inherits the ability to process data online. As a particular choice of approximating family, we blur each basis point with a Gaussian distribution that has a *full covariance matrix* representing the data distribution around that basis point; as a result, we can summarize local data manifold information with a small set of basis points. Our experiments demonstrate that this framework outperforms previous GP classification methods on benchmark datasets in terms of minimizing divergence to the non-sparse GP solution as well as lower misclassification rate.

## 1 Introduction

Gaussian processes (GPs) are powerful nonparametric Bayesian approach to modelling unknown functions. As such, they can be directly used for classification and regression (Rasmussen and Williams, 2006), or embedded into a larger model such as factor analysis (Teh et al., 2005), relational learning (Chu et al., 2006), or reinforcement learning (Deisenroth et al., 2009). Unfortunately, the cost of GPs can be prohibitive for large datasets. Even for the regression case where the GP prediction formula is analytic, training the exact GP model with $N$ points demands an $O(N^3)$ cost for inverting the covariance matrix and predicting a new output requires $O(N^2)$ cost in addition to storing all of the training points.

Ideally, we would like a compact representation, much smaller than the number of training points, of the posterior distribution for the unknown function. This compact representation could be used to summarize training data for Bayesian learning, or it could be passed around as a message in order to do inference in a larger probabilistic model. One successful approach is to compress the training data into a smaller pseudo-dataset over basis points, and then compute the exact posterior that results from the pseudo-dataset. The basis points could literally be a subset of the training instances (Csató, 2002; Lawrence et al., 2002) or they could be artificial "pseudo-inputs" that summarize the training set (Snelson and Ghahramani, 2006). A general framework proposed by Quiñonero-Candela and Rasmussen (2005) shows that many sparse GP regression algorithms, including the "pseudo-inputs" approach, can be viewed as exact inference on an approximate, sparse GP *prior*.

An important advance in sparse GP methods was the insight that the GP model used for inference on the basis points need not match the original GP. For example, the GP applied to the basis points should have a longer length scale than the original, since the data is now sparser (see Walder et al. Walder et al. (2008)). The "variable sigma Gaussian process" (VSGP) algorithm (Walder et al., 2008) allows different length scales for different basis points. Lázaro-Gredilla and Figueiras-Vidal (2009) generalized this idea to allow an arbitrary set of extra parameters (such as a frequency-scale) for each basis point. However, this extension, called Inter-domain Gaussian Processes (IDGPs), as well as VSGP is limited to linear regression.

The reason that previous methods have been limited to linear regression is that they assume that once the pseudo-inputs are chosen, the corresponding pseudo-outputs can be found analytically. For arbitrary likelihoods, not only is it difficult to compute the pseudo-outputs, but it is not even clear what objective the pseudo-outputs should optimize.

In this paper, we provide a new framework, Sparse And Smooth Posterior Approximation (SASPA), for learning sparse GP models with arbitrary likelihoods. Unlike Quiñonero-Candela and Rasmussen (2005)'s work, this new framework is constructed from a different perspective, the *posterior*-approximation perspective: we choose the pseudo-dataset such that the resulting posterior distribution minimizes divergence to the original posterior distribution. This divergence is minimized approximately by expectation propagation (Minka, 2001a). As in (Walder et al., 2008; Lázaro-Gredilla and Figueiras-Vidal, 2009), we allow the GP model on the pseudo-dataset to be different than the original model. However, because the divergence is measured against the posterior from the original model, the modified GP is prevented from overfitting. In the regression case, we obtain exactly the same pseudo-outputs as previous algorithms (SPGP, VSGP, and IDGP). Thus these algorithms can be viewed as special cases of the SASPA framework.

In summary, the main contributions of this paper include the following:

- We present a new divergence-minimization framework, SASPA, for sparse Gaussian process learning. SASPA reduces the computational cost for training GPs from $O(N^3)$ to $O(M^2N)$ where $M$ is the number of basis points.
- In a richer approximating family, we blur each basis point with another distribution. In particular, we use a Gaussian distribution that has a full covariance matrix representing the data distribution around the basis point. Therefore, the SASPA model can effectively summarizes *local data manifold* information with a small set of basis points.
- We describe how to apply the SASPA framework to GP models with regression and classification likelihoods in section 3.2.
- Finally, in section 6, we demonstrate the improved approximation quality of SASPA over previous sparse GP methods on both synthetic data and standard UCI benchmark data.

## 2 Gaussian process models

We denote $N$ independent and identically distributed samples as $\mathcal{D} = \{(\mathbf{x}_1, y_1), \dots, (\mathbf{x}_n, y_n)\}_N$, where $\mathbf{x}_i$ is a $d$ dimensional input and $y_i$ is a scalar output. We assume there is a latent function $f$ that we are modeling and the noisy realization of latent function $f$ at $\mathbf{x}_i$ is $y_i$.

A Gaussian process places a prior distribution over the latent function $f$. Its projection at the samples $\{\mathbf{x}_i\}$ defines a joint Gaussian distribution:

$$p(\mathbf{f}) = \mathcal{N}(\mathbf{f}|\mathbf{m}^0, K)$$

where $m_i^0 = m^0(\mathbf{x}_i)$ is the mean function and $K_{ij} = K(\mathbf{x}_i, \mathbf{x}_j)$ is the covariance function, which encodes the prior notation of smoothness. Normally the mean function is simply set to be zero and we follow this tradition in this paper. A typical kernel covariance function is the squared exponential (known as the Gaussian kernel function):

$$k(\mathbf{x}, \mathbf{x}') = \exp\big(-\frac{||\mathbf{x}' - \mathbf{x}||^2}{2\eta^2}\big), \quad (1)$$

Here $\eta$ is a hyperparameter.

For regression, we use a Gaussian likelihood function

$$p(y_i|f) = \mathcal{N}(y_i|f(\mathbf{x}_i), v_y) \quad (2)$$

where $v_y$ is the observation noise. For classification, the data likelihood has the form

$$p(y_i|f) = (1-\epsilon)\sigma(f(\mathbf{x}_i)y_i) + \epsilon\sigma(-f(\mathbf{x}_i)y_i) \quad (3)$$

where $\epsilon$ models the labeling error and $\sigma(\cdot)$ is a nonlinear function, ie., a cumulative Gaussian distribution or a step function, so that $\sigma(f(x_i)y_i) = 1$ if $f(x_i y_i) \geq 0$ and $\sigma(f(x_i)y_i) = 0$ otherwise.

Given the Gaussian process prior over $f$ and the data likelihood, the posterior process is

$$p(f|\mathcal{D}, \mathbf{t}) \propto GP(f|0, K)\prod_{i=1}^{N} p(y_i|f) \quad (4)$$

Since the Gaussian process is grounded on the $N$ examples, they are called the basis points.

For the regression problem, the posterior process has an analytical form. But to make a prediction on a new sample, we need to invert a $N$ by $N$ matrix. If the training set is big, this matrix inversion will be too costly. For classification or other nonlinear problems, the computational cost is even higher since we do not have a analytical solution to the posterior process and the complexity of the process grows with the number of training samples.

## 3 Sparse-posterior GP: a SASPA perspective

In this section we present the SASPA framework.

### 3.1 Approximating family

We want to compress the training set of $N$ points into a pseudo-dataset of $M$ points, consisting of pseudo-inputs $B = (b_1, \dots, b_M)$, pseudo-outputs $\mathbf{u} = (u_1, \dots, u_M)$, and an $M \times M$ noise covariance matrix $\mathbf{\Lambda}$. The pseudo-dataset will always be a regression dataset, even if the original problem was not regression. The role of the pseudo-dataset is simply to parameterize an approximate posterior distribution on the function $f$. In this sparse representation, we can change the number of points $M$ to regulate its model complexity. Normally, we set $M \ll N$.

Starting from a Gaussian process with covariance function $K$ and zero mean function, the posterior distribution resulting from observing the pseudo-dataset is:

$$q(f) \propto GP(f|0, K)\mathcal{N}(\mathbf{u}|\mathbf{g}_B(f), \mathbf{\Lambda}^{-1}) \quad (5)$$

$$\mathbf{g}_B(f) = \left[\int f(\mathbf{x})\phi(\mathbf{x}|b_1)\mathrm{d}\mathbf{x}, \dots, \int f(\mathbf{x})\phi(\mathbf{x}|b_M)\mathrm{d}\mathbf{x}\right]^{\mathrm{T}}$$

Note that in (6) we have blurred each pseudo-input $b_k$ by a convolving function $\phi(\mathbf{x}|b_k)$, which we will later take to be a Gaussian distribution, representing how the data are distributed *locally* around $b_k$.

The approximate posterior process $q(f)$ is a Gaussian process, so it has a mean and covariance function. To determine these, note that $\mathbf{g}_B(f)$ has a Gaussian *prior* distribution with mean $\mathbf{0}$ and variance $\hat{\mathbf{K}}$:

$$\hat{K}_{ij} = \iint \phi(\mathbf{x}|b_i)K(\mathbf{x}, \mathbf{x}')\phi(\mathbf{x}'|b_j)\mathrm{d}\mathbf{x}\mathrm{d}\mathbf{x}' \quad (6)$$

**Proposition 1** *The posterior process $q(f)$ defined in* (5) *has the mean function $m(\mathbf{x})$ and covariance function $V(\mathbf{x}, \mathbf{x}')$:*

$$m(\mathbf{x}) = \tilde{K}(\mathbf{x}, B)\alpha \quad (7)$$

$$V(\mathbf{x}, \mathbf{x}') = K(\mathbf{x}, \mathbf{x}') - \tilde{K}(\mathbf{x}, B)\beta\tilde{K}(B, \mathbf{x}') \quad (8)$$

*where $\beta = (\hat{\mathbf{K}} + \mathbf{\Lambda}^{-1})^{-1}$, $\alpha = \beta\mathbf{u}$, $\tilde{K}(\mathbf{x}, B) = [\tilde{K}(\mathbf{x}, b_1), \dots, \tilde{K}(\mathbf{x}, b_M)]$, $\tilde{K}(\mathbf{x}, b_j) = \int K(\mathbf{x}, \mathbf{x}')\phi(\mathbf{x}'|b_j)\mathrm{d}\mathbf{x}'$, and $\tilde{K}(B, \mathbf{x}) = (\tilde{K}(\mathbf{x}, B))^{\mathrm{T}}$.*

When using the Gaussian kernel function (1) and setting $\phi(\mathbf{x}|b_i) = \mathcal{N}(\mathbf{x}|a_i, c_i)$ where $b_i = (a_i, c_i)$ , we have

$$\tilde{K}(x, B) = (2\pi\eta^2)^{M/2} \cdot \\ \cdot [\mathcal{N}(\mathbf{x}|a_1, c_1 + \eta^2\mathbf{I}), \dots, \mathcal{N}(\mathbf{x}|a_M, c_M + \eta^2\mathbf{I})]. \quad (9)$$

Due to the space limitation, we omit the proof in this paper (If interested, please see it in the supplemental materials).

To compute the mean and the covariance of the *posterior* distribution of $\mathbf{g}_B(f)$ given the data, we need to convolve the mean and covariance of $q(f)$ with $\phi(\mathbf{x}|b_i)$. Using Proposition 1, we have the following corollary:

**Corollary 2** *The mean and the covariance of the* posterior *distribution of $\mathbf{g}_B(f)$ have the following form:*

$$\tilde{\mathbf{m}}_B = \hat{\mathbf{K}}\alpha \quad (10)$$

$$\tilde{\mathbf{V}}_B = \hat{\mathbf{K}} - \hat{\mathbf{K}}\beta\hat{\mathbf{K}}^{\mathrm{T}} \quad (11)$$

Note that when $\alpha$ and $\beta$ are both zero, $\tilde{\mathbf{m}}_B$ and $\tilde{\mathbf{V}}_B$ reduces to 0 and $\hat{\mathbf{K}}$ for the prior distribution.

### 3.2 Inference by expectation propagation

Assuming $\mathbf{b}$ is given, the remaining question is how to estimate $(\mathbf{u}, \mathbf{\Lambda})$ – or equivalently $(\alpha, \beta)$ – for the sparse posterior process $q(f)$, such that it well approximates the exact posterior (4). We apply expectation propagation to this problem. This is possible because the approximating family in (5) is an exponential family. Each of the original likelihoods $p(y_i|f)$ will be approximated by a message $\tilde{t}_i(f)$ in this family. EP repeats three steps: message deletion, projection, and message update, on each training point. In the message deletion step, we compute the partial belief $q^{\setminus i}(f; \alpha^{\setminus i}, \beta^{\setminus i})$ by removing a message $\tilde{t}_i$ (from the *i-th* point) from the approximate posterior $q(f; \alpha, \beta)$: $q^{\setminus i}(f; \alpha^{\setminus i}, \beta^{\setminus i}) \propto q(f; \alpha, \beta)/\tilde{t}_i$. In the data projection step, we minimize the KL divergence between $\tilde{p}(f) \propto p(y_i|f)q(f; \alpha^{\setminus i}, \beta^{\setminus i})$ and the new approximate posterior $q(f; \alpha, \beta)$, such that the information from each data point is incorporated into the model. Finally, the message $\tilde{t}_i$ is updated based on the new and old posteriors: $\tilde{t}_i \propto q(f; \alpha, \beta)/q^{\setminus i}(f; \alpha^{\setminus i}, \beta^{\setminus i})$.

We start by describing the projection step, since it is the most crucial step in EP. Based on (5), the sparse GP is an exponential family with features $(\mathbf{g}_B(f), \mathbf{g}_B(f)\mathbf{g}_B(f)^{\mathrm{T}})$. As a result, we can determine the sparse GP that minimizes $KL(\tilde{p}(f)|q(f))$ by matching the moments $\tilde{\mathbf{m}}_B$ and $\tilde{\mathbf{V}}_B$ on $\mathbf{g}_B(f)$. The moment matching equations are

$$\tilde{\mathbf{m}}_B = \tilde{\mathbf{m}}_B^{\setminus i} + \tilde{V}^{\setminus i}(B, \mathbf{x}_i)\frac{\mathrm{d}\log Z}{\mathrm{d}m^{\setminus i}(\mathbf{x}_i)} \quad (12)$$

$$\tilde{\mathbf{V}}_B = \tilde{\mathbf{V}}_B^{\setminus i} + \tilde{V}^{\setminus i}(B, \mathbf{x}_i)\frac{\mathrm{d}\log Z}{\mathrm{d}m^{\setminus i}(\mathbf{x}_i)}\tilde{V}^{\setminus i}(\mathbf{x}_i, B) \quad (13)$$

where

$$\tilde{V}^{\setminus i}(B, \mathbf{x})_j = \int \phi(\mathbf{x}'|b_j)V^{\setminus i}(\mathbf{x}', \mathbf{x})\mathrm{d}\mathbf{x}' \quad (14)$$

$$Z = \int q^{\setminus i}(f)p(y_i|f)df \quad (15)$$

and $m^{\setminus i}(\mathbf{x}_i)$ is the mean of $q^{\setminus i}(f)$ at $\mathbf{x}_i$, and $\tilde{V}^{\setminus i}(\mathbf{x}_i, B)$ is covariance matrix of the blurred GP at $\mathbf{x}_i$ and $B$.

Note that these moment matching updates only give the moments of $q(f)$ grounded on $B$. From Proposition 1, the posterior process $q(f)$ is defined by $\alpha$ and $\beta$. So it is sufficient to only update $\alpha$ and $\beta$. To do so, we combine (10)

and (12) and obtain

$$\alpha = \alpha^{\backslash i} + \mathbf{h}\frac{\mathrm{d}\log Z}{\mathrm{d}m^{\backslash i}(\mathbf{x}_i)} \tag{16}$$

where

$$\mathbf{p}_i \triangleq \hat{\mathbf{K}}^{-1}\tilde{K}(B,\mathbf{x}_i)$$
$$\mathbf{h} \triangleq \hat{\mathbf{K}}^{-1}\tilde{V}^{\backslash i}(B,\mathbf{x}_i) = \mathbf{p}_i - \beta^{\backslash i}\tilde{K}(B,\mathbf{x}_i)$$

To obtain the last equation, we apply (8) with $\mathbf{x} = B$ and $\mathbf{x}' = \mathbf{x}_i$. Inserting (11) to (13), we get

$$\beta = \beta^{\backslash i} - \mathbf{h}\mathbf{h}^{\mathrm{T}}\frac{\mathrm{d}^2\log Z}{(\mathrm{d}m^{\backslash i}(\mathbf{x}_i))^2} \tag{17}$$

The above equations define the projection step to obtain the new $q(f)$. Note that in general $KL(\tilde{p}(f)|q(f)) \neq 0$ unless the likelihood function is a Gaussian (e.g., the regression likelihood). For a regression likelihood, the KL is zero and the new $q(f)$ is exact – just like assumed density filters reduce to Kalman filtering.

Given the new $q(f)$, we will update the message $\tilde{t}_i(f)$ according to the ratio $q(f)/q^{\backslash i}(f)$. This leads to the following equations:

$$\tilde{t}_i(f) = \mathcal{N}(\sum_j p_{ij}\int f(\mathbf{x})\phi(\mathbf{x}|b_j)\mathrm{d}\mathbf{x}|g_i, \tau_i^{-1}) \tag{18}$$

where

$$\tau_i^{-1} \triangleq (-\nabla_m^2 \log Z)^{-1} - \tilde{K}(\mathbf{x}_i, B)\mathbf{h} \tag{19}$$
$$g_i \triangleq m^{\backslash i}(\mathbf{x}_i) + (-\nabla_m^2 \log Z)^{-1}\nabla_m \log Z \tag{20}$$

The function $\tilde{t}_i(f)$ can be viewed as a message from the $i\text{-}th$ data point to the sparse GP. To check the validity of this update, note that

$$\tilde{Z} = \int \tilde{t}_i(f)q^{\backslash i}(f)df \propto \mathcal{N}(u_i|\mathbf{p}_i^{\mathrm{T}}\tilde{\mathbf{m}}_B^{\backslash i}, \tau_i^{-1} + \mathbf{p}_i^{\mathrm{T}}\tilde{\mathbf{V}}_B^{\backslash i}\mathbf{p}_i)$$
$$= \mathcal{N}(u_i|m^{\backslash i}(\mathbf{x}_i), \tau_i^{-1} + \tilde{K}(\mathbf{x}_i, B)\mathbf{h}) \tag{21}$$

has the same derivatives as the original $Z = \int t_i(f)q^{\backslash i}(f)df$. Therefore, the multiplication $\tilde{t}_i(f)q^{\backslash i}(f)$ leads to the same $q(f;m,v)$. In other words, $\tilde{t}_i(f) \propto q(f)/q^{\backslash i}(f)$.

To delete a message $\tilde{t}_i(f)$, we need to compute $q^{\backslash i}(f) \propto q(f)/\tilde{t}_i(f)$. Instead of computing this ratio directly, we can equivalently multiply its reciprocal with the current $q(f)$. Then we can solve the multiplication by minimizing $KL(q(f)|q^{\backslash i}(f)\tilde{t}_i(f))$ over $q^{\backslash i}(f)$. Since $q(f)$, $q^{\backslash i}(f)$, and $\tilde{t}_i(f)$ all have the form of the exponential family, this minimal value of this KL is 0, so that $q^{\backslash i}(f) \propto q(f)/\tilde{t}_i(f)$. This KL minimization can be easily done by the moment matching equations very similar to (16) and (17):

$$\alpha^{\backslash i} = \alpha + \mathbf{h}^{\backslash i}\frac{\mathrm{d}\log \tilde{Z}_d}{\mathrm{d}m(\mathbf{x}_i)} \tag{22}$$
$$\beta^{\backslash i} = \beta - \mathbf{h}^{\backslash i}\frac{\mathrm{d}^2\log \tilde{Z}_d}{(\mathrm{d}m(\mathbf{x}_i))^2}(\mathbf{h}^{\backslash i})^{\mathrm{T}} \tag{23}$$

where

$$\mathbf{h}^{\backslash i} \triangleq \mathbf{p}_i - \beta\tilde{K}(B,\mathbf{x}_i) \tag{24}$$
$$\tilde{Z}_d = \int \frac{1}{\tilde{t}_i(f)}q^{\backslash i}(f)\mathrm{d}f$$
$$\propto \mathcal{N}(u_i|\mathbf{p}_i^{\mathrm{T}}\tilde{\mathbf{m}}_B^{\backslash i}, -\tau_i^{-1} + \tilde{K}(\mathbf{x}_i, B)\mathbf{h}^{\backslash i})$$
$$\frac{\mathrm{d}^2\log \tilde{Z}_d}{(\mathrm{d}m(\mathbf{x}_i))^2} = -(-\tau_i^{-1} + \tilde{K}(\mathbf{x}_i, B)\mathbf{h}^{\backslash i})^{-1} \tag{25}$$
$$\frac{\mathrm{d}\log \tilde{Z}_d}{\mathrm{d}m(\mathbf{x}_i)} = (-\frac{\mathrm{d}^2\log Z}{(\mathrm{d}m(\mathbf{x}_i))^2})(g_i - \tilde{K}(\mathbf{x}_i, B)\alpha) \tag{26}$$

Since $\frac{\mathrm{d}^2\log Z}{(\mathrm{d}m^{\backslash i}(\mathbf{x}_i))^2}$ is a scalar and $\mathbf{h}$ is a $M$ by 1 vector, it takes $O(M^2)$ to update $\beta$ via (17). Similarly, it takes $O(M^2)$ to update $\beta^{\backslash i}$ via (23). Therefore, given $N$ training points, the computational cost of SASPA is $O(M^2N)$ per EP iteration over all the data points. Since in practice the number of EP iterations is fairly small, e.g., 10, the overall cost of SASPA is $O(M^2N)$. The EP inference for SASPA is summarized in Algorithm 1.

**Algorithm 1** SASPA

**1.** Initialize $q(f)$, $g_i$, and $\tau_i$ all to be 0.
**2.** Loop until the change over all $g_i$, and $\tau_i$ is smaller than a threshold
  Loop over all training data point $\mathbf{x}_i$
    **Deletion.** Compute $\alpha^{\backslash i}$ and $\beta^{\backslash i}$ for $q^{\backslash i}(f)$ via (22) and (23).
    **Projection.** Compute $\alpha$ and $\beta$ for the posterior $\mathbf{q}(f)$ via (16) and (17).
    **Inclusion.** Update $g_i$, and $\tau_i$ for the message $\tilde{t}_i$ via (19) and (20).

### 3.3 Regression

Given the linear regression likelihood (2), the quantities in the projection step (16)(17) are

$$\frac{\mathrm{d}\log Z}{\mathrm{d}m(\mathbf{x}_i)} = \frac{y_i - m^{\backslash i}(x_i)}{v_y + v^{\backslash i}(x_i, x_i)} \tag{27}$$
$$\frac{\mathrm{d}^2\log Z}{(\mathrm{d}m(\mathbf{x}_i))^2} = \frac{-1}{v_y + v^{\backslash i}(x_i, x_i)} \tag{28}$$

As mentioned before, for the regression case, the KL minimization in the projection step does not introduce approxi-mation error; the approximation aspect of the sparse posterior process $q(f)$ only comes from the compact representation (i.e., fewer basis points than the training points).

### 3.4 Classification

Given the classification likelihood (3) where $\sigma(\cdot)$ is the step function, the quantities in the projection step (16)(17) are

$$z = \frac{m^{\setminus i}(x_i) y_i}{\sqrt{v^{\setminus i}(x_i, x_i)}} \tag{29}$$

$$Z = \epsilon + (1 - 2\epsilon)\psi(z) \tag{30}$$

$$\frac{\mathrm{d}\log Z}{\mathrm{d}m(\mathbf{x}_i)} = \gamma y_i \tag{31}$$

$$\frac{\mathrm{d}^2 \log Z}{(\mathrm{d}m(\mathbf{x}_i))^2} = -\frac{\gamma(m^{\setminus i}(x_i) y_i + v^{\setminus i}(x_i, x_i)\gamma)}{v^{\setminus i}(x_i, x_i)} \tag{32}$$

where $\gamma = \frac{(1-2\epsilon)\mathcal{N}(z|0,1)}{Z\sqrt{v^{\setminus i}(\mathbf{x}_i)}}$ and $\psi(\cdot)$ is the standard Gaussian cumulative distribution function.

## 4 Related work

One of the simplest and fastest approaches to reducing the cost of GPs is to train on a subset of the data. For example, the Informative Vector Machine (IVM) trains on an intelligently chosen subset (Lawrence et al., 2002). Alternatively, we can train on all points, but approximate the contribution of each point. Quiñonero-Candela and Rasmussen (2005) compared several such approximations for regression problems and showed that they can be interpreted as exact inference on an approximate model. This perspective allowed them to show that the SPGP approximation was an improvement over the Deterministic Training Conditional (DTC) approximation. Walder et al. (2008) and Lázaro-Gredilla and Figueiras-Vidal (2009) later extended SPGP to allow basis-dependent length-scales or frequency-scales, and showed that these extensions can also be viewed as exact inference on an approximate model.

However, while the perspective of Quiñonero-Candela and Rasmussen (2005) provides insights for comparing sparse approximations, it is certain limitations as a framework for designing algorithms. First, if we treat the inducing points as model parameters and train them to maximize likelihood, then the approximation may overfit and diverge from the original GP (Titsias, 2009). Second, because the framework relies on exact inference, it only applies to regression problems with linear-Gaussian likelihoods. For classification problems, the inference stage must also be approximate, leaving us with two separate stages of approximation (see e.g. Naish-Guzman and Holden (2007)). Thirdly, this stagewise design is an obstacle to online learning, where we want to interleave the choice of inducing points with the acquisition of new data.

Our work is inspired by the work of Csató and Opper (2000), who showed that online GP classification using EP could be made sparse, using an approximation equivalent to the Fully Independent Training Conditional (FITC) approximation (Note that for regression problems, FITC is the same as SPGP). However, the FITC approximation was introduced as a subroutine, not presented as a part of EP itself. (Naish-Guzman and Holden (2007) presented an equivalent batch algorithm, in which FITC is applied to the GP prior, followed by EP to approximate the likelihood terms.) If we use linear Gaussian likelihoods and the delta function as the blurring function, SASPA reduces to FITC/SPGP for regression. Similarly, with linear Gaussian likelihoods and sphere Gaussians as the blurring functions, SASPA reduces to VSGP.

Csató (2002) later gave a batch EP algorithm, where a different sparse GP approximation (DTC) was used as a subroutine within EP. The software distributed by Csato can run either online or batch and has an option to use either FITC or DTC. In this paper, we re-interpret Csató and Opper (2000)'s work by showing that the FITC approximation can in fact be viewed as part of the overall EP approximation. Furthermore we extend it to include basis-dependent length-scales as in Walder et al. (2008) and Lázaro-Gredilla and Figueiras-Vidal (2009).

An important feature of our method is the use of local manifold information in approximating the exact posterior distribution. To use the valuable local data distribution information, it is possible to first model the data input by a mixture of Gaussians, treat these Gaussians components as basis functions, and then train a Bayesian classifier based on these basis functions. This will be very similar to the Relevance Vector Machine (except the sparse prior part). Unfortunately, this model will not give correct predictive uncertainties, which is well-documented in the case of the RVM (Rasmussen and Quiñonero-Candela, 2005; Quiñonero-Candela et al., 2007). By contrast, SASPA, as a sparse GP model, offers valuable uncertainty quantification for predictions.

Regarding the choice of EP versus other methods, Kuss and Rasmussen (2005) has shown that EP is an excellent choice for GP classification models; EP gives results very close to expensive MCMC both in terms of predictive distributions and marginal likelihood estimates. Also, the EP perspective (not a Variational Bayes treatment) allows us to include several previous approaches as special cases.

## 5 Model selection

The SASPA framework we have presented work for any choice of pseudo-inputs, and does not specify how they should be chosen. Ideally, the pseudo-inputs would be chosen to minimize KL divergence, just like the pseudo-outputs. However, it is not clear how to do this efficiently.

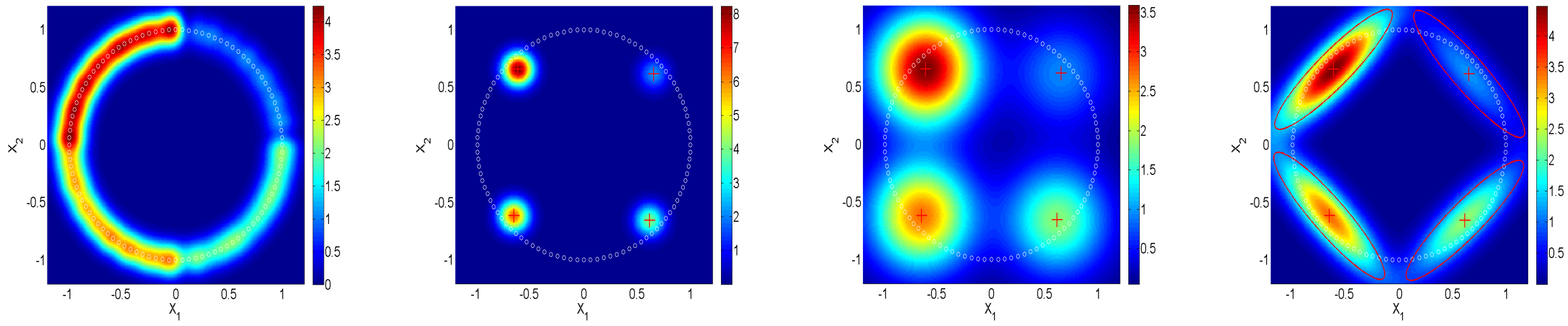

(a). Full GP (b). SPGP (c). $\text{SASPA}_{\text{sphere}}$ (d). $\text{SASPA}_{\text{full}}$

Figure 1: Illustration on simple circle data. The heatmaps represent the values of the posterior means of different methods. Red ellipses and crosses are the mean and the standard deviation of local covariances for SASPA. The white dots are the training data points. $SASPA_{full}$ uses the full local covariance matrix in (d), significantly improving the approximation quality along the circle.

In the literature, one can find a variety of approximate methods for choosing pseudo-inputs. However, for general likelihoods, the only methods currently available are evidence maximization and the IVM. Given so many hyperparameters (basis centers and local covariance matrices), evidence maximization could suffer from overfitting. Actually learning only basis centers for FITC via evidence maximization already suffers from overfitting as discussed in Snelson and Ghahramani (2006). In addition, as a difficult optimization problem, evidence optimization for tuning basis locations is often computationally very expensive. The variational approach of Titsias (2009) does not overfit and is in the same spirit as SASPA, but it minimizes a different divergence measure, and it is only for regression. The IVM uses entropy reduction to select basis points. We did explore the idea of entropy reduction, but we found that in practice it was inferior to a simple efficient clustering algorithm, K-means, for selecting basis points for SASPA. Specifically, we ran K-means on the training inputs, ignoring the outputs, and used the cluster means as pseudo-inputs and the cluster covariances as local covariances. This is similar to the online clustering approach used by Csató (2002).

## 6 Experiments

We evaluate SASPA on both synthetic and real world data and compare its predictive performance with alternative sparse GP methods. Since this paper is specifically about making GPs tractable, we do not compare GP methods against non-GP methods (this has already been done in other papers).

We use the Gaussian kernel for all the experiments. For all of the sparse GP models, we use K-means to choose the basis points (defined as the cluster means) and the covariance matrix $c_i$ (defined as the cluster covariance).

We first examine the performance of the new method on a toy regression problem. Since we can control the generative process of the synthetic data, it is easier for us to gain insight into how the method performs. For regression, we sample 100 data points along a circle with some additive Gaussian noise. The output in different quadrant has different values plus certain additive Gaussian noise.

The mean of the exact posterior distribution of the GP is shown in figure 1a, with approximations in the other panels. Each approximation used the same four basis points, chosen via K-means clustering. Note that when the local covariance matrices become sphere matrices, our method reduces to the multiscale method of (Walder et al., 2008). As shown in the figure, the use of the full local covariance matrix improves the approximation quality.

The next test is a synthetic classification task. Each class is sampled from a multivariate Gaussian distribution. Figure 2 shows an example dataset, with the decision boundaries and the basis points used by each algorithm. Full-GP uses all the training data samples as the basis points in an EP approximation, as described in (Minka, 2001b). FITC-EP uses FITC approximation in an EP framework (Naish-Guzman and Holden, 2007; Csató and Opper, 2000), implemented as a special case of SASPA with no blurring. Finally we have SASPA with sphere and full local covariance matrices for the blurring function $\phi(\mathbf{x}|b_i)$. Quantitative results are shown in figure 3, where we repeatedly sample 200 points for training and 2000 for testing. Sparse Online Gaussian Process (SOGP) (Csató, 2002) corresponds to the DTC approximation applied to the same basis points as the other algorithms. The basis points are chosen by K-means in each case; further improvement is possible by optimizing these. The KL divergence to Full-GP was computed as $\sum_i \mathrm{KL}(p(y_i|x_i) \mid\mid q(y_i|x_i))$ where $x_i$ is a test point and $p(y_i|x_i)$ is the predictive distribution from Full-GP and $q(y_i|x_i)$ is the predictive distribution from the sparse algorithm.

Finally we tested our methods on two standard UCI datasets, Ionosphere and Spambase. The results are summarized in figures 4 and 5. Here we include the Informative

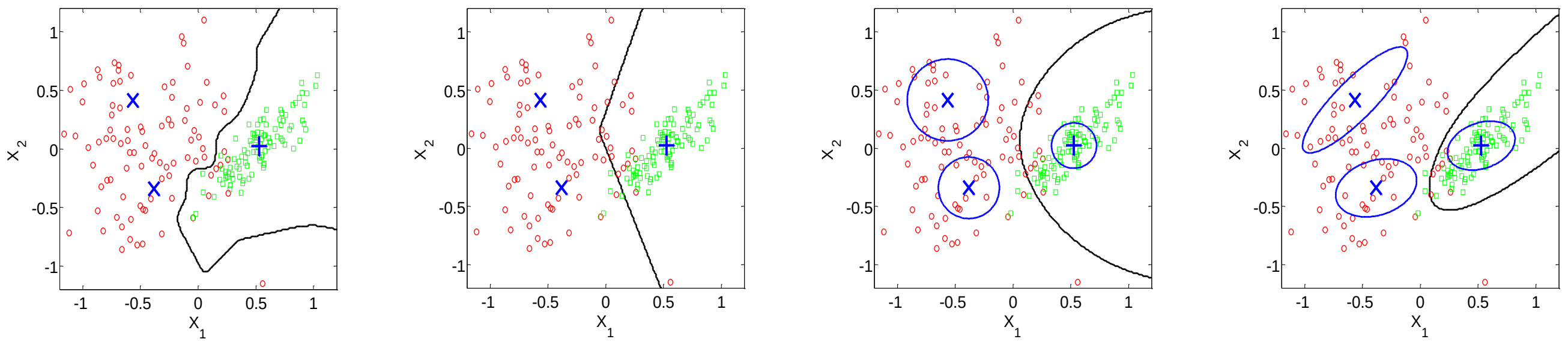

(a). Full GP-EP (b). FITC-EP (c). SASPA$_{\text{sphere}}$ (d). SASPA$_{\text{full}}$

Figure 2: Classification on synthetic data. The blue and red ellipses show the standard deviation of local covariances for SASPA. The black curve is the decision boundary. With only three basis points, the true, complex decision boundary in (a) is well approximated by an ellipse by our method (d).

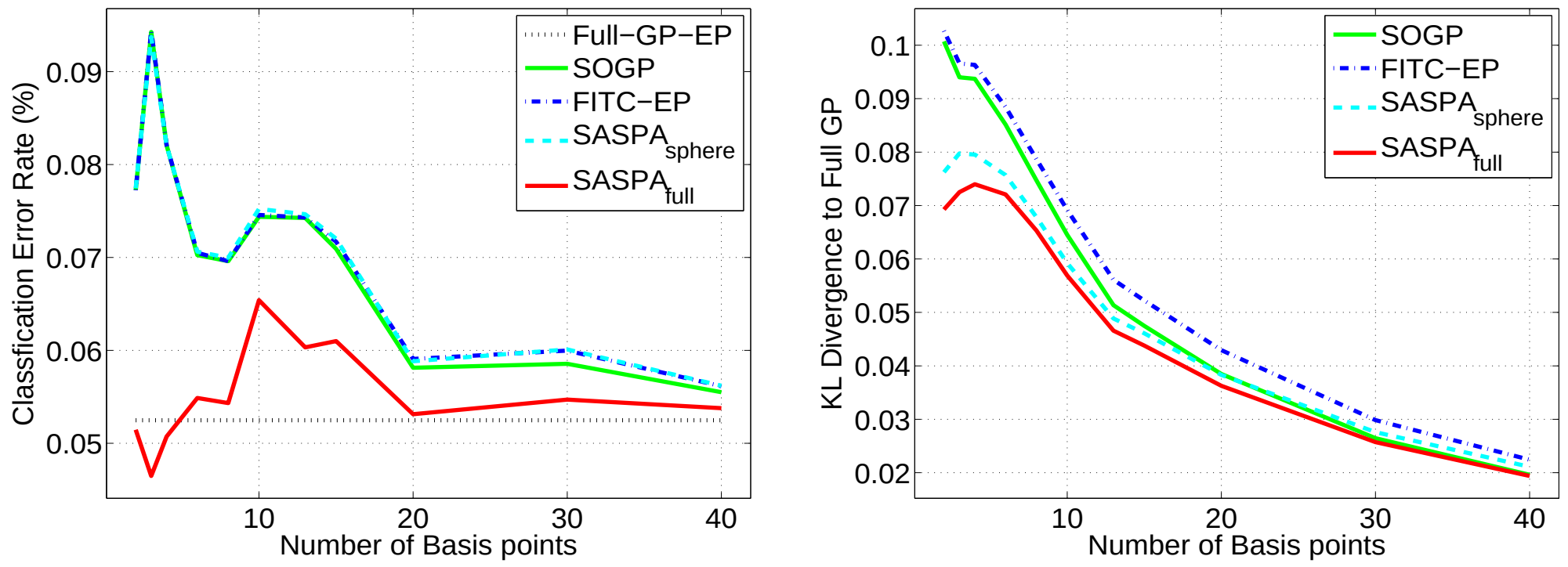

(a). Classification error (b). Approximation error

Figure 3: Effect of different approximation families in SASPA. The test error rates and the approximation error measured by the KL divergence are both averaged over 20 random datasets. Note that all sparse GP algorithms used the same basis point locations. Thus we are emphasizing how well each algorithm makes use of its basis points.

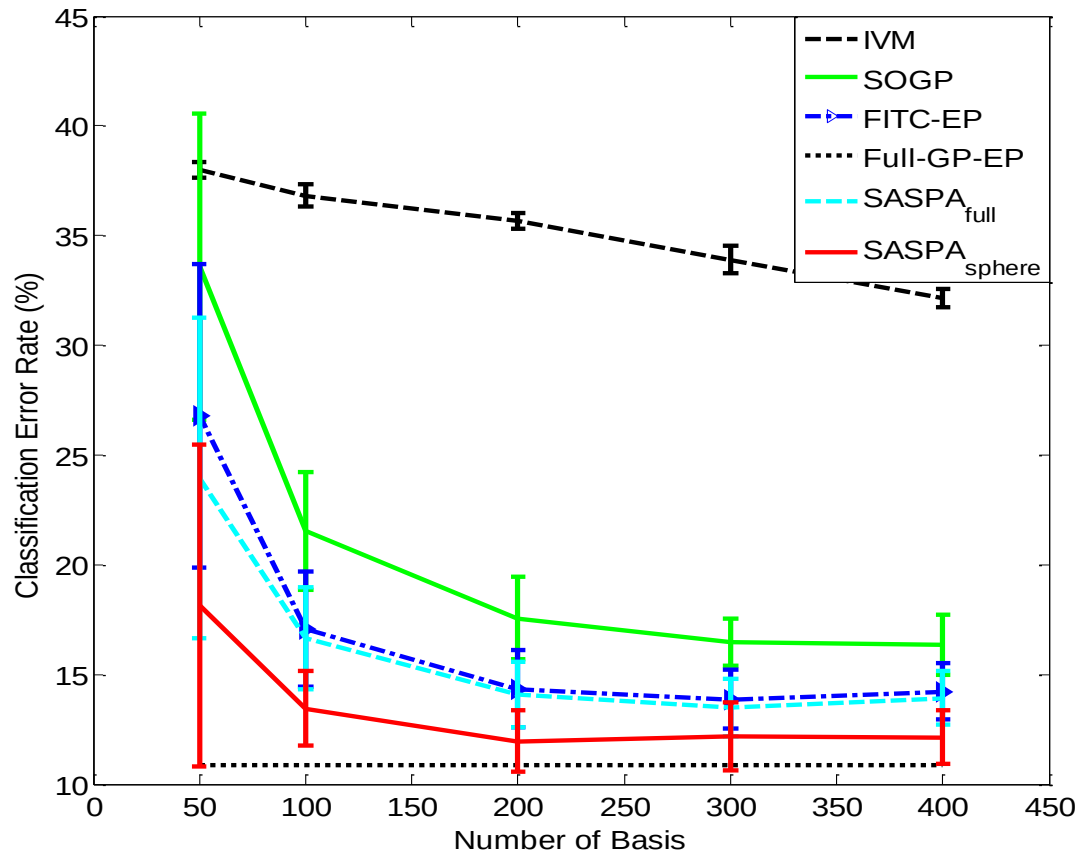

Figure 4: Classification on UCI benchmark dataset Spambase. The results are averaged over 10 random splits of the dataset. For each split, there are 2761 training and 1840 test points. Note that all sparse GP algorithms, except IVM, used the same basis point locations. Thus we are emphasizing how well each algorithm makes use of its basis points.

Relevance Machine (IVM) (Lawrence et al., 2002), which applies one iteration of EP to a subset of the data. We see that basis points chosen by K-means are competitive with the ones chosen by IVM.

## 7 Conclusions

In this paper, we have presented a new sparse GP framework, SASPA, which includes previous dominant sparse GP methods (SPGP, FITC, VSGP, and IDGP) as special cases. Using this framework, we have derived the EP inference methods for general likelihoods. Experimental results demonstrate improved approximation quality and prediction accuracy by enlarging the approximation family to explore local manifold information.

Although empirically the simple clustering approach works well for learning the hyper-parameters, including basis centers and covariances, it remains an interesting open problem to learn these hyper-parameters in a principled and yet tractable way for sparse GPs with general likelihoods.

### Acknowledgements

The work was partially supported by NSF IIS-0916443 and NSF

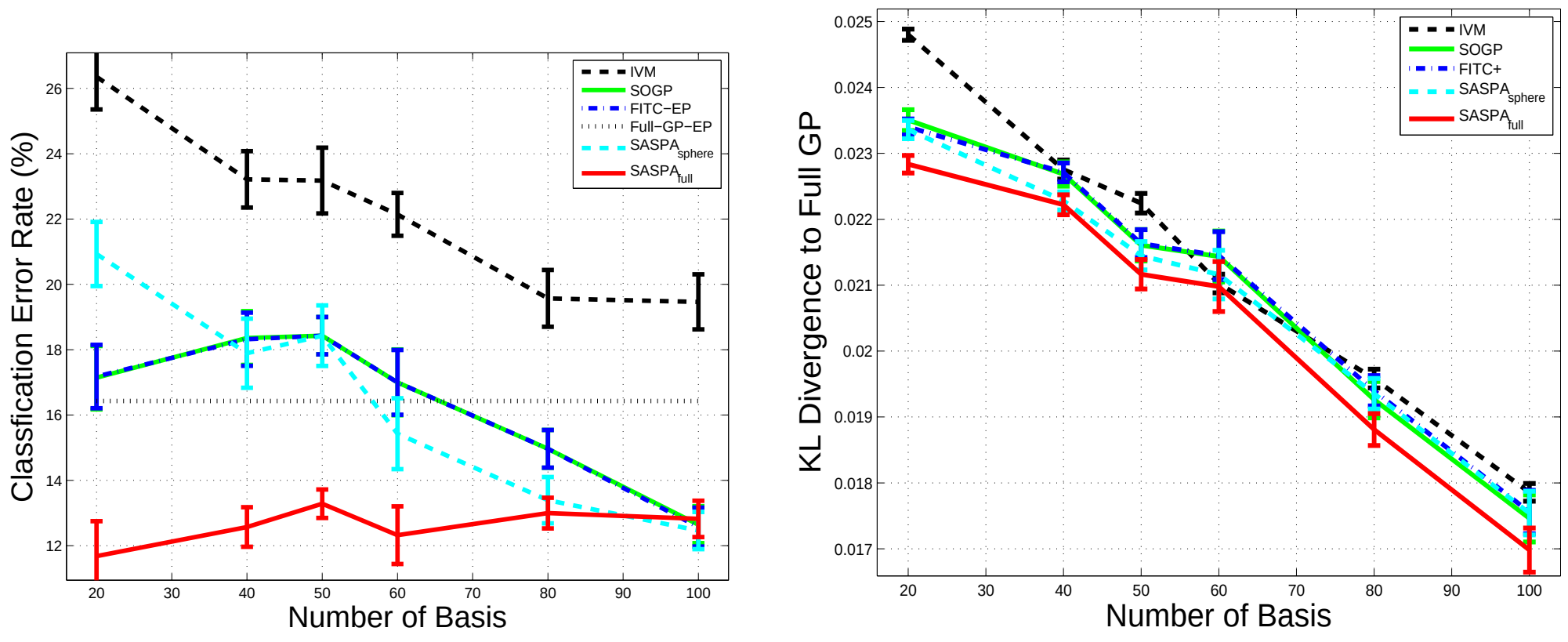

(a). Classification error (b). Approximation error

Figure 5: Classification on UCI dataset Ionosphere. The test error rates and the posterior approximation error are both averaged over 20 random splits. Note that all sparse GP algorithms, except IVM, used the same basis point locations. Note that the relatively small difference in posterior approximation error in (b) translates to the large difference in classification error rates in (a).

## References

Snelson E, Ghahramani Z (2006) Sparse Gaussian processes using pseudo-inputs. In Advances in Neural Information Processing Systems 18. MIT press, pp. 1257–1264.

Walder C, Kim KI, Schölkopf B (2008) Sparse multiscale Gaussian process regression. In ICML '08: Proceedings of the 25th international conference on Machine learning.

Rasmussen CE, Williams CKI (2006) Gaussian Processes for Machine Learning. The MIT Press.

Teh YW, Seeger M, Jordan MI (2005) Semiparametric latent factor models. In The 8th Conference on Artificial Intelligence and Statistics (AISTATS).

Chu W, Sindwhani S, Ghahramani Z, Keerthi SS (2006) Relational learning with Gaussian processes. In Advances in Neural Information Processing Systems 18.

Deisenroth MP, Rasmussen CE, Peters J (2009) Gaussian process dynamic programming. Neurocomputing 72: 1508–1524.

Csató L (2002) Gaussian Processes - Iterative Sparse Approximations. Ph.D. thesis, Aston University.

Lawrence N, Seeger M, Herbrich R (2002) Fast sparse Gaussian process methods: The informative vector machine. In Advances in Neural Information Processing Systems 15. MIT Press, pp. 609–616.

Quiñonero-Candela J, Rasmussen CE (2005) A unifying view of sparse approximate Gaussian process regression. Journal of Machine Learning Research 6: 1935–1959.

Lázaro-Gredilla M, Figueiras-Vidal A (2009) Inter-domain Gaussian processes for sparse inference using inducing features. In Advances in Neural Information Processing Systems 21.

Minka TP (2001a) Expectation propagation for approximate Bayesian inference. In Proceedings of the 17th Conference in Uncertainty in Artificial Intelligence. pp. 362–369.

Titsias MK (2009) Variational learning of inducing variables in sparse Gaussian processes. In Proceedings of the Twelfth International Conference on Artificial Intelligence and Statistics.

Naish-Guzman A, Holden S (2007) The generalized FITC approximation. In Advances in Neural Information Processing Systems 19.

Csató L, Opper M (2000) Sparse representation for Gaussian process models. In Advances in Neural Information Processing Systems 13. MIT Press, pp. 444–450.

Rasmussen CE, Quiñonero-Candela J (2005) Healing the relevance vector machine through augmentation. In Proceedings of the 22nd International Conference on Machine Learning.

Quiñonero-Candela J, Snelson E, Williams O (2007) Sensible priors for sparse Bayesian learning. Technical Report MSR-TR-2007-121, Microsoft Research.

Kuss M, Rasmussen C (2005) Assessing approximate inference for binary Gaussian process classification. Journal of Machine Learning Research 6: 1679–1704.

Minka TP (2001b) A family of algorithms for approximate Bayesian inference. Ph.D. thesis, Massachusetts Institute of Technology.